\documentclass[11pt]{article}
\usepackage{template/coling2020}
\usepackage{times}
\usepackage{url}
\usepackage{latexsym}

\usepackage{xcolor}
\usepackage{soul}
\usepackage{breakcites}
\usepackage{natbib}
\usepackage{tabularx}
\usepackage{caption}
\usepackage{booktabs}
\usepackage{multirow}
\usepackage{hyperref}
\usepackage{enumitem}
\usepackage{amsmath}
\usepackage{amsfonts}
\usepackage{float}

\colingfinalcopy 

\title{Continual Lifelong Learning in Natural Language Processing: A Survey}
\author{
    Magdalena Biesialska
    \qquad
    Katarzyna Biesialska
    \qquad
    Marta R. Costa-juss\`a \\
    Universitat Polit\`ecnica de Catalunya, Barcelona, Spain \\
    \texttt{\{magdalena.biesialska,katarzyna.biesialska,marta.ruiz\}@upc.edu}
}
\date{}

\begin{document}
\maketitle
\begin{abstract}
 Continual learning (CL) aims to enable information systems to learn from a continuous data stream across time. However, it is difficult for existing deep learning architectures to learn a new task without largely forgetting previously acquired knowledge. Furthermore, CL is particularly challenging for language learning, as natural language is ambiguous: it is discrete, compositional, and its meaning is context-dependent. In this work, we look at the problem of CL through the lens of various NLP tasks. Our survey discusses major challenges in CL and current methods applied in neural network models. We also provide a critical review of the existing CL evaluation methods and datasets in NLP. Finally, we present our outlook on future research directions.
\end{abstract}

\section{Introduction}
\label{sec:intro}
Human beings learn by building on their memories and applying past knowledge to understand new concepts. Unlike humans, existing neural networks (NNs) mostly learn in isolation and can be used effectively only for a limited time. Models become less accurate over time, for instance, due to the changing distribution of data -- the phenomenon known as \textit{concept drift} \citep{schlimmer1986incremental,widmer1993effective}. With the advent of deep learning, the problem of continual learning (CL) in Natural Language Processing (NLP) is becoming even more pressing, as current approaches are not able to effectively retain previously learned knowledge and adapt to new information at the same time. 

Throughout the years, numerous methods have been proposed to address the challenge known as \textit{catastrophic forgetting} (CF) or \textit{catastrophic interference} \citep{mccloskey1989catastrophic}. Naïve approaches to mitigate the problem, such as retraining the model from scratch to adapt to a new task (or a new data distribution), are costly and time-consuming. This is reinforced by the problems of \textit{capacity saturation} and \textit{model expansion}. Concretely, a parametric model, while learning data samples with different distributions or progressing through a sequence of tasks, eventually reaches a point at which no more knowledge can be stored -- i.e. its representational capacity approaches the limit \citep{sodhani2018training,aljundi2018selfless}. At this point, either model's capacity is expanded, or a selective forgetting -- which likely incurs performance degradation -- is applied. The latter choice may result either in a deterioration of prediction accuracy on new tasks (or data distributions) or forgetting the knowledge acquired before. This constraint is underpinned by a defining characteristic of CL, known as the \textit{stability-plasticity dilemma}. Specifically, the phenomenon considers the model's attempt to strike a balance between its \textit{stability} (the ability to retain prior knowledge) and its \textit{plasticity} (the ability to adapt to new knowledge).

CL in the NLP domain, as opposed to computer vision or robotics, is still nascent \citep{greco-etal-2019-psycholinguistics,sun2020LAMOL}. The differences are reflected in the small number of proposed methods aiming to alleviate the aforementioned issues and the evaluation benchmarks. To the best of our knowledge, apart from the work of \citet{chen2018lifelong}, our paper is the only study summarizing the research progress related to continual, lifelong learning in NLP.

%
%
\blfootnote{
    %
    %
    \hspace{-0.65cm}  
    This work is licensed under a Creative Commons 
    Attribution 4.0 International License.
    License details:
    \url{http://creativecommons.org/licenses/by/4.0/}.
    %
    %
    %
    %
}

\section{Learning Paradigms}
\label{sec:learning-paradigms}
In this section, we discuss principles of CL and related machine learning (ML) paradigms, as well as contemporary approaches to mitigate CF.

\subsection{Continual Learning}
\label{sec:cl}

Continual learning\footnote{Continual learning in the literature is also referred to as: \textit{lifelong learning} \citep{silver2002task,silver2013lifelong,chen2018lifelong,chaudhry2019efficient,parisi2019continual,aljundi2017expert}, \textit{incremental learning} \citep{solomonoff1989system,chaudhry2018riemannian}, \textit{sequential learning} \citep{mccloskey1989catastrophic,shin2017continual,aljundi2018selfless}, \textit{explanation-based learning} \citep{thrun1996learning}, and \textit{never-ending learning} \citep{carlson2010toward}.} \citep{Ring1994ContinualLI} is a machine learning paradigm, whose objective is to adaptively learn across time by leveraging previously learned tasks to improve generalization for future tasks. Hence, CL studies the problem of sequential learning from a continuous stream of data, drawn from a potentially non-stationary distribution, and reusing gained knowledge throughout the lifetime while avoiding CF. 

More formally: the goal is to sequentially learn a model $f: \mathcal{X} \times \mathcal{T} \rightarrow \mathcal{Y}$ from a large number of tasks ${\mathcal{T}}$. The model is trained on examples ($x_{i}, y_{i}$), such that: $x_{i} \in \mathcal{X}_{t_{i}}$ is an input feature vector, $y_{i} \in \mathcal{Y}_{t_{i}}$ is a target vector (e.g. a class label), and $t_{i} \in \mathcal{T}$ denotes a task descriptor (in the simplest case $t_{i}=i$) where $i \in \mathbb{Z}$. The objective is to maximize the function $f$ (parameterized by $\theta \in \mathbb{R}$) at the task $\mathcal{T}_{i}$, while minimizing CF for tasks $\mathcal{T}_{1}\dots\mathcal{T}_{i-1}$.

Although the above-mentioned definitions of CL may seem fairly general, there are certain desired properties, which are summarized in Table \ref{table:desiderata}.

\begin{table}[!ht]
\footnotesize
\centering
\def\arraystretch{1}
\begin{tabularx}{\textwidth}{ll}
\toprule
\multicolumn{1}{c}{\textit{Property}} & \multicolumn{1}{c}{\textit{Definition}} \\
\toprule
\addlinespace[0.6em]
\textbf{Knowledge retention} & The model is not prone to catastrophic forgetting. \\
\addlinespace[0.3em]
\textbf{Forward transfer} & The model learns a new task while reusing knowledge acquired from previous tasks. \\
\addlinespace[0.3em]
\textbf{Backward transfer} & The model achieves improved performance on previous tasks after learning a new task. \\
\addlinespace[0.3em]
\textbf{On-line learning} & The model learns from a continuous data stream.\\
\addlinespace[0.3em]
\textbf{No task boundaries} & The model learns without requiring neither clear task nor data boundaries.\\
\addlinespace[0.3em]
\textbf{Fixed model capacity} & Memory size is constant regardless of the number of tasks and the length of a data stream. \\
\addlinespace[0.2em]
\bottomrule
\end{tabularx}
\caption{Desiderata of continual learning.}
\label{table:desiderata}
\end{table}

In practice, current CL systems often relax at least one of the requirements listed in Table \ref{table:desiderata}. Most methods still follow the off-line learning paradigm -- models are trained using batches of data shuffled in such a way as to satisfy the independent and identically distributed (i.i.d.) assumption. Consequently, many models are trained solely in a supervised fashion with large labeled datasets, and thus they are not exposed to more challenging situations involving few-shot, unsupervised, or self-supervised learning. Additionally, existing approaches often fail to restrict themselves to make a single pass over the data, and this entails longer learning times. Moreover, the number of tasks as well as their identity are frequently known to the system from the outset.

\subsection{Related Machine Learning Paradigms}
\label{sec:related-ml-paradigms}

Traditionally, many ML models are designed to be trained for merely a single task. However, it has been proven that transferring knowledge learned from one task and applying it to another task is a powerful mechanism for NNs. In many respects, CL bears some resemblance to other dominant learning approaches. Therefore, in this section, we draw connections between various ML paradigms. We provide an overview of the approaches, and in particular, we shed light on the shared principles as well as on the aspects that make CL different from other ML paradigms (see Table \ref{table:related-paradigms}).

\begin{table}[!h]
\footnotesize
\centering
\def\arraystretch{0.1}
\begin{tabularx}{\textwidth}{llll}
\toprule
\textit{Paradigm} & \multicolumn{1}{c}{\textit{Definition}}  & \multicolumn{1}{c}{\textit{Properties*}} & \multicolumn{1}{c}{\textit{Related works}} \\
\toprule
    \parbox{1.6cm}{\textbf{Transfer\\learning}} & 
    \parbox{4.9cm}{Transferring knowledge from a source task/domain to a target task/domain to improve the performance of the target task.} &
    \parbox{4.0cm}{
        \begin{itemize}[noitemsep,topsep=0pt,leftmargin=*,parsep=0pt,partopsep=0pt,labelindent=1em,labelsep=0.1cm]
        \item[+] forward transfer
        \item[--] no backward transfer
        \item[--] no knowledge retention
        \item[--] task boundaries
        \item[--] off-line learning
        \end{itemize}} &
    \scriptsize\parbox{3.8cm}{\begin{flushleft}
        \citep{pan2010transfer} \citep{howard-ruder-2018-universal} \citep{peters-etal-2018-deep} \citep{radford2019language} \citep{devlin-etal-2019-bert} \citep{houlsby2019parameter} \citep{raffel2019exploring}\end{flushleft}}\\
\midrule
    \parbox{1.6cm}{\textbf{Multi-task\\learning}} & 
    \parbox{4.9cm}{Learning multiple related tasks jointly, using parameter sharing, to improve the generalization of all the tasks.} & 
    \parbox{4.0cm}{
        \begin{itemize}[noitemsep,topsep=0pt,leftmargin=*,parsep=0pt,partopsep=0pt,labelindent=1em,labelsep=0.1cm]
        \item[+] positive transfer
        \item[--] negative transfer
        \item[--] task boundaries
        \item[--] off-line learning
        \end{itemize}} &  
    \scriptsize\parbox{3.8cm}{\begin{flushleft}\citep{Caruana1997} \citep{zhang2017survey} \citep{ruder2017overview} \citep{McCann2018decaNLP} \citep{stickland19a} \citep{phang2020english}\end{flushleft}}\\
\midrule
    \parbox{1.6cm}{\textbf{Meta-learning}} & 
    \parbox{4.9cm}{\textit{Learning to learn.} Learning generic knowledge, given a small set of training examples and numerous tasks, and quickly adapting to a new task.} &
    \parbox{4.0cm}{
        \begin{itemize}[noitemsep,topsep=0pt,leftmargin=*,parsep=0pt,partopsep=0pt,labelindent=1em,labelsep=0.1cm]
        \item[+] forward transfer
        \item[--] no backward transfer
        \item[--] no knowledge retention
        \item[--] off-line learning
        \end{itemize}} &
    \scriptsize\parbox{3.8cm}{ 
        \begin{flushleft}\citep{thrun-pratt-1998} \citep{finn2017} \citep{xu-ijcai2018} \citep{obamuyide-vlachos-2019-meta} \citep{hospedales2020metalearning} \citep{beaulieu2020learning}\end{flushleft}}\\
\midrule
    \parbox{1.6cm}{\textbf{Curriculum\\learning}} & 
    \parbox{4.9cm}{Learning from training examples arranged in a meaningful order -- task or data difficulty gradually increases.} &
    \parbox{4.0cm}{
    \begin{itemize}[noitemsep,topsep=0pt,leftmargin=*,parsep=0pt,partopsep=0pt,labelindent=1em,labelsep=0.1cm]
        \item[+] forward transfer
        \item[+] backward transfer
        \item[+] knowledge retention    
        \item[--] task boundaries
        \item[--] off-line learning
    \end{itemize}} &
    \scriptsize\parbox{3.8cm}{\begin{flushleft}\citep{elman1993learning} \citep{bengio2009curriculum} \citep{van-der-wees-etal-2017-dynamic}\\\citep{zhang2018empirical} \citep{zhang-etal-2019-curriculum} \citep{platanios-etal-2019-competence} \citep{ruiter2020selfinduced}\end{flushleft}}\\
\midrule
    \parbox{1.6cm}{\textbf{On-line\\learning}} &
    \parbox{4.9cm}{Learning over a continuous stream of training examples provided in a sequential order. Experiences \textit{concept drift} due to non-i.i.d. data.} &
    \parbox{4.0cm}{
        \begin{itemize}[noitemsep,topsep=0pt,leftmargin=*,parsep=0pt,partopsep=0pt,labelindent=1em,labelsep=0.1cm]
        \item[+] on-line learning
        \item[+] forward transfer
        \item[--] no backward transfer
        \item[--] no knowledge retention
        \item[--] single task/domain
        \end{itemize}} &
    \scriptsize\parbox{3.8cm}{ 
        \begin{flushleft}\citep{Bottou-1999} \citep{Bottou-LeCun-2004} \citep{cesa-bianchi_lugosi_2006} \citep{Shalev-Shwartz-2012} \citep{c-de-souza-etal-2015-online} \citep{hoi2018online}\end{flushleft}}\\
\midrule
    \parbox{1.6cm}{\textbf{On-the-job\\learning}} &
    \parbox{4.9cm}{Discovering new tasks, learning and adapting on-the-fly. On-the-job learning operates in an \textit{open-world} environment, and it involves interaction with humans and the environment. It belongs to the CL family of methods.} &
    \parbox{4.0cm}{
        \begin{itemize}[noitemsep,topsep=0pt,leftmargin=*,parsep=0pt,partopsep=0pt,labelindent=1em,labelsep=0.1cm]
        \item[+] on-line learning
        \item[+] forward transfer
        \item[+] backward transfer
        \item[+] knowledge retention
        \item[+] no task boundaries
        \item[+] open-world learning
        \item[--] interactive learning
        \end{itemize}} &
    \scriptsize\parbox{3.8cm}{ 
        \begin{flushleft} \citep{xu2019open-world} \\\citep{mazumder-etal-2019-lifelong} \\\citep{liu2020learning-on-the-job}\end{flushleft}}\\
\bottomrule
\end{tabularx}
\caption{Comparison of related ML paradigms. \textit{*}
Properties aligned (+) and unaligned (--) with CL.}
\label{table:related-paradigms}
\end{table}

In principle, we assume that the ability of a model to generalize can be considered one of its most important characteristics. Importantly, if tasks are related, then knowledge transfer between tasks should lead to a better generalization and faster learning \citep{lopez2017gradient,sodhani2018training}. Therefore, we compare the paradigms taking into account how well they are able to leverage an inductive bias. Specifically, positive backward transfer improves the performance on old tasks, while negative backward transfer deteriorates the performance on previous tasks (if high, it enables CF). Similarly, negative forward transfer impedes learning of new concepts, while positive forward transfer allows to learn a new task with just a few examples (if high, it enables zero-shot learning).

\subsection{Approaches to Continual Learning}
\label{sec:cl-approaches}
The majority of existing CL approaches tend to apply a single model structure to all tasks \citep{pmlr-v97-li19m} and control CF by various scheduling schemes. We distinguish three main families of methods: \textit{rehearsal}, \textit{regularization}, and \textit{architectural} as well as a few hybrid categories. Importantly, the number of models originating purely from the NLP domain is quite limited. 

\paragraph{Rehearsal methods} rely on retaining some training examples from prior tasks, so that they can later be shown to a task at hand. \citet{rebuffi2017icarl} proposed the most well-known method for incremental class learning, i.e. the iCaRL model. Furthermore, as training samples are kept per each task and are periodically replayed while learning the model, the computing and memory requirements of the model increase proportionally to the number of tasks. To reduce storage, it is advised to use either latent replay \citep{pellegrini2019latent} or pseudo-rehearsal \citep{robins1995catastrophic} methods.

\paragraph{Pseudo-rehearsal methods} 
are a sub-group of rehearsal methods. Instead of using training samples from memory, pseudo-rehearsal models generate examples by knowing the probability distributions of previous task samples. Notable approaches include a generative autoencoder \citep[FearNet,][]{kemker2018fearnet} and a model based on Generative Adversarial Networks \citep[DGR,][]{shin2017continual}.

\paragraph{Regularization methods} are single-model approaches that rely on a fixed model capacity with an additional loss term that aids knowledge consolidation while learning subsequent tasks or data distributions. 
For instance, Elastic Weight Consolidation \citep[EWC,][]{kirkpatrick2016overcoming} reduces forgetting by regularizing the loss; in other words, it slows down the learning of parameters important for previous tasks.

\paragraph{Memory methods} are a special case of regularization methods that can be divided into two groups: \textit{synaptic regularization} \citep{zenke2017continual,kirkpatrick2016overcoming,chaudhry2018riemannian} and \textit{episodic memory} \citep{lihoiem2016learning,jung2016less,lopez2017gradient,Chaudhry2019ContinualLW,dautume2019episodic}. The former methods are focused on reducing interference with the consolidated knowledge by adjusting learning rates in a way that prevents changes to previously learned model parameters. While the latter store training samples from previously seen data, which are later rehearsed to allow learning new classes.
Importantly, Gradient Episodic Memory \citep[GEM,][]{lopez2017gradient} allows positive backward transfer and prevents the loss on past tasks from increasing. Other notable examples of this approach include A-GEM \citep{chaudhry2019efficient}, MER \citep{riemer2018learning}, or a method originating from NLP - MBPA++ \citep{dautume2019episodic}.

\paragraph{Knowledge distillation methods} bear a close resemblance to \textit{episodic memory} methods, but unlike GEM they keep the predictions at past tasks invariant \citep{rebuffi2017icarl,lopez2017gradient}. In particular, it is a class of methods alleviating CF by relying on knowledge transfer from a large network model (teacher) to a new, smaller network (student) \citep{hinton-distill}. The underlying idea is that the student model learns to generate predictions of the teacher model.
As demonstrated in \citet{kim2016sequence} and \citet{wei-etal-2019-online}, knowledge distillation approaches can prove especially suitable for neural machine translation models, which are mostly large, and hence reduction in size is beneficial.

\paragraph{Architectural methods} prevent forgetting by applying modular changes to the network's architecture and introducing task-specific parameters. Typically, previous task parameters are kept fixed \citep{rusu2016progressive,mancini2018adding} or masked out \citep{serr2018overcoming,Mallya2018PackNetAM}. Moreover, new layers are often injected dynamically to augment a model with additional modules to accommodate new tasks.
Progressive Networks \citep[PNN,][]{rusu2016progressive}, and their improved versions: Dynamically Expandable Network \citep[DEN,][]{yoon2018lifelong}, Reinforced Continual Learning \citep[RCL,][]{xu2018reinforced}, are prominent examples. The main drawback of such strategies is the substantially growing number of parameters. Similar to PNN, BatchEnsemble \citep{Wen2020BatchEnsemble} is also immune to CF, in addition it supports parallel order of tasks and consumes less memory than PNN thanks to training only fast weights. In a similar vein, adapter modules aim to overcome the problem of a large number of parameters. They act as additional network layers with a small number of parameters \citep{rebuffi2017learning,houlsby2019parameter,bapna-firat-2019-simple,pfeiffer2020AdapterHub} that reconfigure the original network on-the-fly for a target task, while keeping the parameters of the original network untouched and shared between different tasks.

\section{Evaluation}
\label{sec:evaluation}
Even though CL is now experiencing a surge in the number of proposed new methods, there is no unified approach when it comes to their evaluation using benchmark datasets and metrics \citep{parisi2019continual}. And as we will show in this section, this is especially true in the NLP domain. There is a scarcity of datasets and benchmark evaluation schemes available specifically for CL in NLP.

\subsection{Protocols}
\label{sec:protocols}

Basically, researchers often focus on evaluating the \textit{plasticity} (generalization) side and the \textit{stability} (consistency) side of the model.
Various protocols and methodologies for CL method evaluation have been devised throughout the years \citep[e.g.][]{Kemker2017MeasuringCF,serr2018overcoming,sodhani2018training,pflb2019comprehensive,chaudhry2019efficient}; however, many of them suffer from deficiencies such as small datasets or a limited number of evaluated methods, to name a few.

Furthermore, as observed by \citet{chaudhry2019efficient}, the prevalent learning protocol followed in many CL research efforts stems from supervised learning, where many passes over the data of each task are performed. The authors claimed that in a CL setting this approach is flawed as with more passes over the data of a given task, the model degrades more because it forgets previously acquired knowledge. In a similar vein, \citet{yogatama2019learning} contended that NLP models are predominantly evaluated with respect to their performance on a held-out test set, which is measured after the training is done for a given task. Therefore, \citet{chaudhry2019efficient} introduced a learning protocol that, according to the authors, is more suitable for CL as it satisfies the constraint of a single pass over the data, which is motivated by the need for a faster learning process. Another recent approach, proposed by \citet{dautume2019episodic}, relies on a sequentially presented stream of examples derived from various datasets in one pass, without revealing dataset boundary or identity to the model.

\subsection{Benchmarks and Metrics}
\label{sec:benchmarks-metrics}

For years the NLP domain has lagged behind computer vision and other ML areas \citep[e.g.][]{kirkpatrick2016overcoming,zenke2017continual,lomonaco2017core50,rebuffi2017icarl} when it comes to the availability of standard CL-related benchmarks \citep{greco-etal-2019-psycholinguistics,wang-etal-2019-sentence}. However, the situation has slightly improved recently with an introduction of a handful of multi-task benchmarks. In particular, \textsc{GLUE} \citep{wang-etal-2018-glue,greco-etal-2019-psycholinguistics} and \textsc{SuperGLUE} \citep{wang2019superglue} benchmarks track performance on eleven and ten language understanding tasks respectively, using existing NLP datasets. Along the same line, \citet{McCann2018decaNLP} presented the Natural Language Decathlon (\textsc{decaNLP}) benchmark for evaluating the performance of models across ten NLP tasks.
The decathlon score (decaScore) is an additive combination of various metrics specific for each of the ten selected tasks (i.e. the normalized F1 metric, BLEU and ROUGE scores, among others).
Similar to \textsc{decaNLP}, a recently proposed Cross-lingual TRansfer Evaluation of Multilingual Encoders (\textsc{XTREME}) benchmark \citep{hu2020xtreme} also uses a diverse set of NLP tasks and task-specific measures to evaluate the performance of cross-lingual transfer learning. \textsc{XTREME} consists of nine tasks derived from four different categories and uses zero-shot cross-lingual transfer with the English language as the source language for evaluation. 

In principle, CL models should not only be evaluated against traditional performance metrics (such as model accuracy); it is also important to measure their ability to reuse prior knowledge. Similarly, evaluating how quickly models learn new tasks is also essential in the CL setting. Although CF is crucial to address in CL systems, there is no consensus on how to measure it \citep{pflb2019comprehensive}. Arguably, the two most popular and general metrics to address this issue are \textit{Average Accuracy} and \textit{Forgetting Measure} \citep{lopez2017gradient,chaudhry2018riemannian,Chaudhry2019ContinualLW}. The former evaluates the average accuracy, while the latter measures forgetting after the model is trained continually on all the given task mini-batches. Concretely, we aim to measure test performance on the dataset $\mathcal{D}$ for each of the ${\mathcal{T}}$ tasks, letting $a_{j, i}$ be the performance of the model on the held-out test set of task $t{i}$ after the model is trained on task $t{j}$. Later \citet{chaudhry2019efficient} proposed the third metric, \textit{Learning Curve Area} (\textit{LCA}), that measures how quickly a model is able to learn. The three metrics are defined as follows:

\begin{itemize}
    \item \textbf{Average Accuracy}: $A \in[0,1]$ \citep{chaudhry2018riemannian}. The average accuracy after incremental training from the first task to ${\mathcal{T}}$ is given as:
$$A_{\mathcal{T}}=\frac{1}{\mathcal{T}} \sum_{i=1}^{\mathcal{T}} a_{\mathcal{T}, i}$$
    \item \textbf{Forgetting Measure}: $F \in[-1,1]$ \citep{chaudhry2018riemannian}. The average forgetting measure after incremental training from the first task to ${\mathcal{T}}$ is defined as:
    $$F_{\mathcal{T}}=\frac{1}{\mathcal{T}-1} \sum_{i=1}^{\mathcal{T}-1} f_{i}^{\mathcal{T}}$$
    where $f_{i}^{j}$ is the forgetting on task $t{i}$ after the model is trained up to task $t{j}$ and computed as:
    $$f_{i}^{j}=\max _{k \in\{1, \cdots, j-1\}} a_{k, i}-a_{j, i}$$
    \item \textbf{Learning Curve Area}: $LCA \in[0,1]$ \citep{chaudhry2019efficient}. LCA is the area under the $Z_{b}$ curve, which captures the learner's performance on all ${\mathcal{T}}$ tasks. $Z_{b}$ is the average accuracy after observing the $b$-th mini-batch and is defined as:
    $$Z_{b}=\frac{1}{\mathcal{T}} \sum_{i=1}^{\mathcal{T}} a_{i, b, i}$$
    where $b$ denotes the mini-batch number.
\end{itemize}

Similarly, \citet{Kemker2017MeasuringCF} proposed three metrics for evaluating CF, i.e. the metrics evaluate the ability of a model to retain previously acquired knowledge and how well it acquires new information. In the NLP domain, \citet{yogatama2019learning} introduced a new metric, based on an online (prequential) encoding \citep{blier2018description}, which measures the adoption rate of an existing model to a new task. Specifically, the metric called \textit{online codelength} $\ell(\mathcal{D})$ is defined as follows:
\begin{equation*}
\ell(\mathcal{D})=\log _{2}|y|-\sum_{i=2}^{N} \log _{2} p\left(y_{i} | x_{i} ; \theta_{\mathcal{D}_{i-1}}\right)
\end{equation*}
where $|y|$ denotes the number of possible labels (classes) in the dataset $\mathcal{D}$, and $\theta_{\mathcal{D}_{i}}$ stands for the model parameters trained on a particular subset of the dataset. Similar to \textit{LCA} \citep{chaudhry2019efficient}, \textit{online codelength} is also related to an area under the learning curve.

While most CL methods consider settings without human-in-the-loop, some allow a human domain expert to provide the model with empirical knowledge about the task at hand. For instance, \citet{prokopalo-etal-2020-evaluation} introduced the evaluation of human assisted learning across time by leveraging user-defined model adaptation policies for NLP and speech tasks, such as machine translation and speaker diarization.

\subsection{Evaluation Datasets}
\label{sec:evaluation-datasets}

Most widely adopted CL benchmark datasets are image corpora such as \textsc{Permuted MNIST} \citep{kirkpatrick2016overcoming}, \textsc{CUB-200} \citep{WelinderEtal2010,WahCUB_200_2011}, or split \textsc{CIFAR-10/100} \citep{lopez2017gradient}. Benchmark corpora have also been proposed for objects - \textsc{Core50} \citep{lomonaco2017core50} and sound - \textsc{AudioSet} \citep{Gemmeke2017AudioSA}. However, none of the well-established standard datasets used in the CL field is related to NLP. Therefore, due to the scarcity of NLP-curated datasets, some of the above-mentioned datasets have also been utilized for NLP scenarios.

\begin{table}[!hb]
\footnotesize
\centering
\begin{tabular}{lll}
\toprule
\multicolumn{1}{c}{\textit{Name}} & \multicolumn{1}{c}{\textit{Details}} & \multicolumn{1}{c}{\textit{Related works}}\\
\toprule
    \parbox{4.5cm}{\textbf{\textsc{XCOPA} - Cross-lingual Choice of Plausible Alternatives}} &
    \parbox{8cm}{
        \begin{itemize}[noitemsep,topsep=0pt,leftmargin=*,parsep=0pt,partopsep=0pt,labelindent=0em,labelsep=0.1cm]{
        \item a typologically diverse multilingual dataset for causal commonsense reasoning, which is the translation and reannotation
        \item covers 11 languages from distinct families}
        \end{itemize}} &
    \scriptsize\parbox{2.3cm}{ 
        \begin{flushleft}
        \citep{ponti2020xcopa}
        \end{flushleft}}\\
\midrule
    \parbox{4.5cm}{\textbf{\textsc{WebText}}} &
    \parbox{8cm}{
        \begin{itemize}[noitemsep,topsep=0pt,leftmargin=*,parsep=0pt,partopsep=0pt,labelindent=0em,labelsep=0.1cm]
        \item  a dataset of millions of webpages suitable for learning language models without supervision
        \item 45 million links scraped from Reddit, 40 GB dataset
        \end{itemize}} &
    \scriptsize\parbox{2.3cm}{ 
        \begin{flushleft}
        \citep{radford2019language}
        \end{flushleft}}\\
\midrule
    \parbox{4.5cm}{\textbf{\textsc{C4} - Colossal Clean Crawled Corpus}} &
    \parbox{8cm}{
        \begin{itemize}[noitemsep,topsep=0pt,leftmargin=*,parsep=0pt,partopsep=0pt,labelindent=0em,labelsep=0.1cm]
        \item a dataset constructed from Common Crawl’s web crawl corpus and serves as a source of unlabeled text data
        \item 17 GB dataset
        \end{itemize}} &
    \scriptsize\parbox{2.3cm}{ 
        \begin{flushleft}
        \citep{raffel2019exploring}
        \end{flushleft}}\\
\midrule
    \parbox{4.5cm}{\textbf{\textsc{Lifelong FewRel} - Lifelong Few-Shot Relation Classification Dataset}} &
    \parbox{8cm}{
        \begin{itemize}[noitemsep,topsep=0pt,leftmargin=*,parsep=0pt,partopsep=0pt,labelindent=0em,labelsep=0.1cm]
        \item sentence-relation pairs derived from Wikipedia distributed over 10 disjoint clusters (representing different tasks)
        \end{itemize}} &
    \scriptsize\parbox{2.3cm}{ 
        \begin{flushleft}
        \citep{wang-etal-2019-sentence} \citep{obamuyide-vlachos-2019-meta}
        \end{flushleft}}\\
\midrule
    \parbox{4.5cm}{\textbf{\textsc{Lifelong Simple Questions}}} &
    \parbox{8cm}{
        \begin{itemize}[noitemsep,topsep=0pt,leftmargin=*,parsep=0pt,partopsep=0pt,labelindent=0em,labelsep=0.1cm]
        \item single-relation questions divided into 20 disjoint clusters (i.e. resulting in 20 tasks)
        \end{itemize}} &
    \scriptsize\parbox{2.3cm}{ 
        \begin{flushleft}
        \citep{wang-etal-2019-sentence}
        \end{flushleft}}\\
\bottomrule
\end{tabular}
\caption{An overview of major NLP benchmark datasets to evaluate multi-task and CL methods.}
\label{table:benchmarks}
\end{table}

Similarly, in the absence of NLP benchmark corpora, the majority of papers use adopted versions of popular NLP datasets. One such example is domain adaptation, where researchers frequently use different, standard NLP corpora for in-domain and out-of-domain datasets.
\citet{farquhar2018towards} stressed that prior research often presented incomplete evaluations, and utilized dedicated CL datasets or environments that cannot be considered general, one-size-fits-all benchmarks. As the scholars argued, such benchmarks are useful in narrow cases, limited to their respective subdomains.
\noindent The number of NLP-specific CL datasets is still very limited, even though there have been lately a few notable attempts to create such corpora (summarized in Table {\ref{table:benchmarks}}).

Importantly, as \citet{parisi2019continual} contended, with the increasing complexity of the evaluation dataset at hand, the overall performance of the model often decreases. The scholars attributed this to the fact that the majority of methods are tailored to work only for less complex scenarios, as they are not robust and flexible enough to alleviate CF in less controlled experimental conditions. In a similar vein, \citet{yogatama2019learning} stressed that the recent tendency to construct datasets that are easy to solve without requiring generalization or abstraction is an impediment toward general linguistic intelligence. Hence, we advocate further research on establishing challenging evaluation datasets and evaluation metrics for CL in NLP that will allow to capture how well models generalize to new, unseen tasks.

\section{Continual Learning in NLP Tasks}
\label{sec:nlp-tasks}
Natural language processing covers a diverse assortment of tasks. Despite the variety of NLP tasks and methods, there are some common themes. On a syntax level, sentences in any domain or task follow the same syntax rules. Furthermore, regardless of task or domain, there are words and phrases that have almost the same meaning. Therefore, sharing of syntax and semantic knowledge should be feasible across NLP tasks. In this section, we explore how CL methods are used in most popular NLP tasks.

\subsection{Word and Sentence Representations}
\label{sec:embeddings}

Distributed word vector representations underlie many NLP applications. Although high-quality word embeddings can considerably boost performance in downstream tasks, they cannot be considered a silver bullet as they suffer from inherent limitations. Typically, word embeddings are trained on large-size general corpora, as the size of in-domain corpora is in most cases not sufficient. This comes at a cost, since embeddings trained on general-purpose corpora are often not suitable for domain-specific downstream tasks, and in result, the overall performance suffers. In a CL setting, this also implies that vocabulary may change with respect to two dimensions: time and domain. There is an established consensus that the meaning of words changes over time due to complicated linguistic and social processes \citep[e.g.][]{kutuzov-etal-2018-diachronic,shoemark-etal-2019-room}. Hence, it is important to detect and accommodate shifts in meaning and data distribution, while preventing previously learned representations from CF.

In general, a CL scenario for word and sentence embeddings has not received much attention so far, except for a handful of works. To tackle this problem, for example \citet{xu-ijcai2018} proposed a meta-learning method, which leverages knowledge from past multi-domain corpora to generate improved new domain embeddings. \citet{liu-etal-2019-continual} introduced a sentence encoder updated over time using matrix conceptors to continually learn corpus-dependent features. Importantly, \citet{wang-etal-2019-sentence} argued that when a NN model is trained on a new task, the embedding vector space undergoes undesired changes, and in result the embeddings are infeasible for previous tasks. To mitigate the problem of embedding space distortion, they proposed to align sentence embeddings using anchoring. Recently a research line at the intersection of word embeddings and language modeling, termed contextual embeddings, has emerged and demonstrates state-of-the-art results across numerous NLP tasks. In the next section, we will look closely at how this approach to learning embeddings is geared towards CL.

\subsection{Language Modeling}
\label{sec:lm}

Contextual representations learned via unsupervised pre-trained language models (LMs), such as: \textsc{ULMFiT} \citep{howard-ruder-2018-universal}, \textsc{ELMo} \citep{peters-etal-2018-deep} or \textsc{BERT} \citep{devlin-etal-2019-bert}, allow to attain strong performance on a wide range of supervised NLP tasks. Precisely, thanks to inductive transfer, complex task-specific architectures have become less needed. In consequence, the process of training many neural-based NLP systems boils down to two steps: (1) an NN-based language model is trained on a large unlabeled text data; (2) this pre-trained language representation model is then reused in supervised downstream tasks. In principle, a large LM trained on a sufficiently large and diverse corpus is able to perform well across many datasets and domains \citep{radford2019language}. Furthermore, \citet{gururangan2020dont} studied the effects of task-adaptation as well as domain-adaptation on the transferability of adapted pre-trained LMs across domains and tasks. The authors concluded that continuous domain- and task-adaptive pre-training of LMs leads to performance gains in downstream NLP tasks.

Research interest in LM-based methods for CL in NLP has recently spiked. \citet{dautume2019episodic} proposed an episodic memory-based model, \textsc{MbPA++}, that augments the encoder-decoder architecture. In order to continually learn, \textsc{MbPA++} also performs sparse experience replay and local adaptation. The scholars claimed that \textsc{MbPA++} trains faster than \textsc{A-GEM}, and it does not take longer to train it than an encoder-decoder model. While this is possible due to sparse experience replay, yet \textsc{MbPA++} requires extra memory. In a similar vein, \textsc{LAMOL} \citep{sun2020LAMOL} is based on language modeling. Unlike \textsc{MbPA++}, this method does not use any extra memory. \textsc{LAMOL} mitigates CF by means of pseudo-sample generation, as the model is trained on the mix of new task data and pseudo old samples.

\subsection{Question Answering}
\label{sec:qa}

Question answering (QA) is considered a traditional NLP task, encompassing reading comprehension as well as information and relation extraction among others. Conceptually, it is also very much related to conversational agents, such as chatbots and dialogue agents. Hence, not only in research settings but even more so in real-life scenarios (e.g. in the conversation), it is immensely important for such systems to continuously extract and accumulate new knowledge \citep{chen2018lifelong}. It is believed that a good dialogue agent should be able to not only interact with users by responding and asking questions, but also to learn from both kinds of interaction \citep{li2016learning}.

Although question answering is a stand-alone NLP task, some researchers \citep[e.g.][]{kumar2016ask,McCann2018decaNLP} proposed to view NLP tasks through the lens of QA. In the context of CL, both \citet{dautume2019episodic} and \citet{sun2020LAMOL} reported experimental results on a QA task. Research in dialogue agents, which are able to continually learn, is a very active area  \citep[e.g.][]{Gasic2014IncrementalOA,su2016continuously}. Findings of \citet{li2017DialogueLW} indicate that a conversational model initially trained with fixed data can improve itself, when it learns from interactions with humans in an on-line fashion. Interestingly, information and relation extraction were an early subject of research interest in CL. Information extraction is considered one of the first research areas, which embraced the goal of never-ending learning. A semi-supervised \textsc{NELL} \citep{carlson2010toward} and an unsupervised \textsc{ALICE} \citep{banko2007alice} systems, which iteratively extract information and build general domain knowledge, were at the forefront of CL in NLP. In the case of relation extraction, \citet{wang-etal-2019-sentence} introduced an embedding alignment method to enable CL for relation extraction models. Also, \citet{obamuyide-vlachos-2019-meta} proposed to extend the work of \citet{wang-etal-2019-sentence} by framing the lifelong relation extraction as a meta-learning problem; however, without the costly need for learning additional parameters.

\subsection{Sentiment Analysis and Text Classification}
\label{sec:sa-tc}

Sentiment analysis (SA) is a popular choice for evaluating models on text classification. Arguably the most pressing problem of current approaches to SA is their poor performance on new domains. Therefore, various domain adaptation methods have been proposed to improve the performance of SA models in the multi-domain scenario \citep[consult][]{barnes2018-projecting}. This issue is of utmost importance if one thinks about CL in sentiment classification. One of the earliest approaches to CL for SA was proposed in \citet{chen-etal-2015-lifelong}. According to \citet{chen2018lifelong}, CL can enable SA models to adapt to a large number of domains, since many new domains may already be covered by other past domains. Additionally, SA systems should become more accurate not only in classification but also in the discovery of word polarities in specific domains. Research in opinion about aspects has been conducted as well. \citet{shu-etal-2016-lifelong} presented an unsupervised CL approach to classify opinion targets into entities and aspects. Furthermore, \citet{shu-etal-2017-lifelong} proposed a method based on conditional random fields to improve supervised aspect extraction across time. Experiments on text classification in the CL setting were performed in \citet{dautume2019episodic} and \citet{sun2020LAMOL}.

\subsection{Machine Translation}
\label{sec:mt}
The approach introduced by \citet{Luong-Manning:iwslt15} laid the groundwork for subsequent studies in adapting neural machine translation (NMT). More specifically, the authors explored the adaptation through continued training, where an NMT model trained using large corpora in one domain can later initialize a new NMT model for another domain. Their findings suggested that fine-tuning of the NMT model trained on out-of-domain data using a small in-domain parallel corpus boosts performance. Likewise, other works  \citep[e.g.][]{freitag2016fast,Chu2017AnEC} supported this claim. \citet{khayrallah2018regularized} pointed out that, due to over-fitting, some amount of knowledge learned from the out-of-domain corpus is being forgotten during fine-tuning. Hence, such domain adaptation techniques are prone to CF. NMT models experience difficulties when dealing with data from diverse domains, hence we argue this is not a sufficient solution. As dominant fine-tuning approaches require training and maintaining a separate model for each language or domain, \citet{bapna-firat-2019-simple} proposed to add light-weight task-specific adapter modules to support parameter-efficient adaptation. We further argue that NMT -- as opposed to phrase-based MT -- rarely incorporates translation memory, and so it is inherently harder for NMT models to adapt using active or interactive learning. However, some attempts have been made \citep[e.g.][]{peris-casacuberta-2018-active,liu-etal-2018-learning-actively,kaiser2017,tu-etal-2018-learning}. In a similar vein, there were approaches \citep{sokolov-etal-2017-shared} to incorporate bandit learners, which implicitly involve domain adaptation and on-line learning, for MT systems. We share the viewpoint of \citet{farajian-etal-2017-multi}, that NMT models ultimately should be able to adapt on-the-fly to on-line streams of diverse data (i.e. language, domain), and thus CL for NMT is essential. 

While domain adaptation methods are widely used in the context of adapting NMT models, there have also been other attempts. Multilingual NMT \citep{dong-etal-2015-multi,firat-etal-2016-multi,Ha2016TowardMN,johnson-etal-2017-googles,tan2018multilingual} can be framed as a multi-task learning problem. Multilingual NMT aims to use a single model to translate between multiple languages. Such systems are beneficial not only because they can handle multiple translation directions using a single model, and thus reduce training and maintenance costs, but also due to joint training with high-resource languages they can improve performance on low- and zero-resource languages \citep{arivazhagan2019massively}. To eliminate the need for retraining the entire NMT system, \citet{escolano2020multilingual} proposed a language-specific encoder-decoder architecture, where languages are mapped into a shared space, and either encoder or decoder is frozen when training on a new language.

Another related research line is curriculum learning. Most approaches concentrate on the selection of training samples according to their relevance to the translation task at hand. Different methods have been applied, for example, \citet{van-der-wees-etal-2017-dynamic} and \citet{zhang-etal-2019-curriculum} adapted a model to a domain by introducing samples which are increasingly domain-relevant or domain-distant respectively. Curriculum methods based on difficulty and competence were explored in \citet{zhang2018empirical} and \citet{platanios-etal-2019-competence}. \citet{ruiter2020selfinduced} proposed a self-supervised NMT model, that uses data selection to train on increasingly complex and task-related samples in combination with a denoising curriculum.

A stream of research focused on techniques more traditionally associated with CL has also been present. In the works of  \citet[]{miceli-barone-etal-2017-regularization,khayrallah2018regularized,varis-bojar-2019-unsupervised,thompson2019overcoming} regularization approaches (e.g. EWC) were leveraged. Furthermore, \citet{kim2016sequence} explored knowledge distillation, where the student model learns to match the teacher’s actions at the word- and sequence-level. \citet{wei-etal-2019-online} proposed an on-line knowledge distillation approach, in which the best checkpoints are utilized as the teacher model.  Lately, \citet{Li2020Compositional} demonstrated that label prediction continual learning leveraging compositionality brings improvements in NMT.

\section{Research Gaps and Future Directions}
\label{sec:future-research}
Although there is a growing number of task-specific approaches to CL in NLP, nevertheless, the body of research work remains rather scant \citep{sun2020LAMOL,greco-etal-2019-psycholinguistics}.
While the majority of current NLP methods is task-specific, we believe task-agnostic approaches will become much more prevalent. Contemporary methods are limited along three dimensions: data, model architectures, and hardware.

In the real world, we often deal with partial information data. Moreover, data is drawn from non-i.i.d. distributions, and is subject to agents' interventions or environmental changes. Although attempts exist, where a model learns from a stream of examples without knowing from which dataset and distribution they originate from \citep[e.g.][]{dautume2019episodic}, such approaches are rare. Furthermore, learning on a very few examples (e.g. via few-shot transfer learning) \citep{liu2020learning-on-the-job} is a major challenge for current models, even more so performing out-of-distribution generalization \citep{Bengio_2019}. In particular, widely used in NLP sequence-to-sequence models still struggle with \textit{systematic generalization} \citep{lake2018generalization,bahdanau2018systematic}, being unable to learn general rules and reason about high-level language concepts. For instance, recent work on counterfactual language representations by \citet{feder2020causalm} is a promising step in that direction.
The non-stationary learning problem can be alleviated by understanding and inferring causal relations from data \citep[e.g.][]{osawa2019practical} -- which is an outstanding challenge \citep{pearl2009causality} -- and coming up with combinations that are unlikely to be present in training distributions \citep{Bengio_2019}. Namely, language is compositional; hence, the model can dynamically manipulate the semantic concepts which can be recombined in novel situations \citep{lake2015human} and later supported by language-based \textit{abductive reasoning} \citep[e.g.][]{bhagavatula2020abductive}.

On a model level, a combination of CL with Bayesian principles should allow to identify better the importance of each parameter of an NN and aid parameter pruning and quantization \citep[e.g.][]{Ebrahimi2020Uncertainty-guided,golkar2019continual}.
We believe that not only the parameter informativeness should be uncertainty-guided, but also the periodic replay of previous memories should be informed by causality. 
Furthermore, it is important to focus on reducing model capacity and computing requirements. Even though the over-parametrization of NNs is pervasive \citep{neyshabur2018towards}, many current CL approaches promote the expansion of parameter space.
We envision further research efforts focused on compression methods, such as knowledge distillation, low-rank factorization and model pruning. Importantly, while CL allows for continuous adaptation, we believe that integrating CL with meta-learning has the potential to further unlock generalization capabilities in NLP. As meta-learning is able to efficiently learn with limited samples, hence such a CL model would adapt quicker in dynamic environments \citep[e.g.][]{ritter18been,shedivat2018continuous}. This would be especially beneficial for NLP systems operating in low-resource language and domain settings.

Finally, further research aiming at developing comprehensive benchmarks for CL in NLP would be an important addition to the existing studies. On the one hand, we observe a proliferation of multi-task benchmarks \citep[e.g.][]{McCann2018decaNLP,wang-etal-2018-glue,wang2019superglue}. On the other hand, the CL paradigm and evaluation of CL systems call for more robust approaches than traditional performance metrics (e.g. accuracy, F1 measure) and multi-task evaluation schemes with clearly defined data and task boundaries.

\section{Conclusion}
\label{sec:conclusion}
In this work, we provided a comprehensive overview of existing research on CL in NLP. We presented a classification of ML paradigms and methods for alleviating CF, as well as discussed how they are applied to various NLP tasks. Also, we summarized available benchmark datasets and evaluation approaches. Finally, we identified research gaps and outlined directions for future research endeavors. We hope this survey sparks interest in CL in NLP and inspires to view linguistic intelligence in a more holistic way.

\section*{Acknowledgements}
We would like to thank the anonymous reviewers for their helpful feedback. We also would like to thank Marc'Aurelio Ranzato for providing a detailed clarification of the \textit{LCA} metric. We thank Carlos Escolano for a fruitful discussion on Table \ref{table:related-paradigms}.
This work is supported in part by the Catalan Agencia de Gestión de Ayudas Universitarias y de Investigación (AGAUR) through the FI PhD grant; the Spanish Ministerio de Ciencia e Innovación and by the Agencia Estatal de Investigación through the Ramón y Cajal grant and the project PCIN-2017-079; and by the European Research Council (ERC) under the European Union’s Horizon 2020 research and innovation programme (grant agreement No. 947657).

\newpage

\bibliographystyle{template/acl_natbib}
\bibliography{references}

\end{document}